\documentclass[]{ceurart}
\usepackage{placeins}
\usepackage{rotating} 
\usepackage{float}
\usepackage{listings}
\usepackage{caption}
\usepackage{pdflscape}
\usepackage{enumitem}
\usepackage{changepage}
\begin{document}

\copyrightyear{2025}
\copyrightclause{Copyright for this paper by its authors.
  Use permitted under Creative Commons License Attribution 4.0
  International (CC BY 4.0).}
\cortext[1]{Corresponding author.}  
\conference{The 16th Annual Learning Analytics and Knowledge Conference (LAK) Workshop on LLM Psychometrics, April 27, 2026, Bergen, Norway}
    \title{Correcting Human Labels for Rater Effects in AI Evaluation: An Item Response Theory Approach}

\author[1]{Jodi M. Casabianca}[%
email=jodi@broadmetrics.co,
orcid=0000-0002-1644-6731,
url=https://broadmetrics.co/,
]
\cormark[1]
\fnmark[1]
\address[1]{BroadMetrics,
  New Jersey, USA}
\address[2]{CUNY Graduate Center, New York, USA}

\author[2]{Maggie Beiting-Parrish}[
email=magdalen.beiting@gmail.com,
orcid=0000-0002-3998-8672
]

\begin{abstract}
Human‐provided evaluations play a central role in training and assessing AI models, yet these data are rarely treated as measurements subject to systematic error. This paper integrates psychometric rater models into the AI pipeline to improve the reliability and validity of conclusions drawn from human judgments. The paper reviews common rater effects, severity and centrality, that distort observed ratings, and demonstrates how item response theory rater models, particularly the multi-faceted Rasch model, can separate true output quality from rater behavior. Using the OpenAI summarization dataset as an empirical example, we show how adjusting for rater severity produces corrected estimates of summary quality and provides diagnostic insight into rater performance. Incorporating psychometric modeling into human-in-the-loop evaluation offers more principled and transparent use of human data, enabling developers to make decisions based on adjusted scores rather than raw, error-prone ratings. This perspective highlights a path toward more robust, interpretable, and construct-aligned practices for AI development and evaluation.
\end{abstract}


\maketitle

\section{Introduction}
A growing body of research in natural language processing (NLP) and artificial intelligence (AI) has drawn attention to the instability and unreliability of human evaluation data \cite{fabbri2021summeval,hardy2025all}. These findings parallel a longstanding problem that psychometricians in the educational testing domain have researched, which is how to deal with and account for rater errors in scoring \cite{casabiancapsych}. A plethora of psychometric studies show that rater severity, centrality, and other rater effects complicate the inference of true performance from observed ratings. Leveraging psychometric methods to account for these errors in ratings would provide AI evaluation scientists with tools to obtain refined estimates of AI model quality. The purpose of this paper is to demonstrate a visionary perspective on how to integrate psychometric modeling into the AI system pipeline to account for imperfect human data when evaluating AI models. We demonstrate this approach using an empirical dataset that evaluates summarization models.

\section{The Problem with Human Data}
Human annotations are a cornerstone of NLP/AI research, yet they are fraught with well-documented problems that undermine their reliability as “ground truth.” Unlike automated metrics, human data carries variability that reflects differences in interpretation, attention, and even motivation. 

Human raters frequently disagree with one another when asked to rate the same material using the same scoring guide or rubric \cite{fabbri2021summeval}. This inconsistency reflects poor inter-rater reliability, which limits the replicability of evaluation outcomes. Psychometric frameworks highlight that such inconsistency introduces random error variance, reducing the reliability of scores and obscuring true differences in model quality. Even when raters are \textit{consistent}, they may be \textit{inaccurate} if their judgments do not properly reflect the construct of interest. Inaccuracy leads to misleading conclusions about model performance.

\subsection{Rater Effects}
Rater behavior is understood to introduce systematic effects or patterns in ratings, not just random errors. These \textit{rater effects} can distort observed scores in predictable ways. Understanding these patterns is critical for designing evaluation protocols and statistical models that correct for them. Two frequently measured rater effects are:
\begin{itemize}
    \item \textbf{Severity/Leniency}: \textit{Severe} raters consistently assign lower scores than others, independent of the quality of the response. Conversely, \textit{lenient} raters assign higher scores. In AI evaluation, this manifests when some raters give systematically harsher ratings to machine-generated outputs, which may be mistaken for a real difference in system quality; two models of identical quality may receive different ratings simply because one rater is harsher than another. 
    \item \textbf{Centrality/Extremity}: Raters exhibiting \textit{centrality} have a preference for using only the middle of a rating scale (e.g., using only score points 3, 4, 5 on a 1 - 7 scale). Raters exhibiting \textit{extremity} gravitate toward the extreme ends of the score scale. Centrality biases compress score distributions, obscuring true performance differences, while extremity exaggerates them. For Likert-style evaluations \cite{likert1932technique}, this can mean the difference between data that shows clear distinctions across systems and data that misleadingly suggests uniformity or polarization. 
\end{itemize}

There are several other rater effects including restriction of range and halo effects. See \cite{casabianca2022rater,myford2003part1,myford2004part2} for more background on rater effects.

\subsection{Limitations of Using Human Raters in AI Systems}
In large scale educational testing contexts in which essays, short answers, and spoken responses are scored, companies employ very large pools of human raters who are highly trained and calibrated. Many of these raters are experienced career-raters who are committed long-term to scoring assessments. Some research has shown that large rater effects are not prevalent in these rater pools \cite{nieto2019accounting}. While there can be noisy raters who degrade the reliability of scores, this does not lead to specific error patterns or score shifts because the vast majority of these large rater pools are aligned on a rubric. 

In AI systems, human data is used in different stages of the AI system pipeline: data for fine-tuning, preference data for reinforcement learning, data to train and align LLM judges, and human labeling for quality and other metrics. Many raters or "annotators" are hired from crowdsourcing networks such as Amazon Mechanical Turk (MTurk) or hired by annotation or "data-as-a-service" companies, who then onboard subject matter experts for temporary projects. Unlike large testing organizations, the human rater pools in these instances may not be that large, well-trained, or motivated \cite{maline2025unmasking}, indicating that if rater effects are prevalent in only a few raters, they may be impacting the labels in a consequential way. 

Minimizing the error and aberrant rater tendencies in the annotation system is the first line of defense against noisy or error-laden labels. It is best practice to minimize rater errors by design of the system (see Section 4 of \cite{mccaffrey2022best} for a detailed discussion or \cite{mcclellan2010constructed} for an overview). Unfortunately, there is a large cost to designing and managing high quality scoring systems. Thus, the main focus of this paper is not on minimizing the error from raters by design, but on how we can adjust for rater aberrance after-the-fact via psychometric modeling.

\section{Estimating and Mitigating Rater Effects in Rating Scale Data}
Item response theory (IRT) models are probabilistic models used in testing to place individuals (test takers) on a latent trait scale based on their responses to test items \cite{lord2012applications}. The latent trait in educational testing might be mathematical knowledge or writing skill. In AI systems, it is AI-relevant constructs or traits such as coherence, quality, trustworthiness, etc., that are being measured. These are constructs we cannot observe and measure directly. In their simplest form, IRT models generally only account for item and test taker characteristics, however, explanatory IRT models can be specified with additional parameters. For example, the class of IRT rater models \cite{casabianca2022rater,robitzsch2018item,wolfe2012application} incorporate parameters for raters to adjust for these effects in the trait estimates. Models such as this become useful in annotation systems in which there are $N$ samples or outputs that are being labeled by a human either for the purposes of creating training data or evaluation.

\subsection{Multi-facet Rasch Modeling for Rater Effects Estimation}
A very popular and widely used IRT rater model is the polytomous Many-Facet Rasch Model (MFRM) \cite{linacre1989many}, which extends the Rasch framework by incorporating additional facets such as raters. The standard MFRM specifies the log-odds of an output receiving a particular rating category on a given dimension (e.g., quality). In our setting, an \emph{output} ($n$=1,...,$N$) refers to a single AI-generated response or output (e.g., a summary or answer). Each output is evaluated on one or more \emph{items} ($i$=1,...,$I$), where an item corresponds to a rating dimension such as factual accuracy, relevance, or coherence. Human judges or \emph{raters} ($j$=1,...,$J$) assign ordinal ratings using a predefined rating scale (e.g., 1 = poor, 2 = fair, 3 = good, 4 = excellent). A \emph{rating category} ($k$=1,...,$K$) therefore corresponds to a specific score level on this scale. Under this formulation, the MFRM specifies the log-odds of output $n$ receiving category $k$ (as opposed to $k-1$) on item $i$ from rater $j$ as
\[
\log \left( \frac{P_{nijk}}{P_{nij(k-1)}} \right)
= \theta_n - \delta_i - \rho_j - \tau_{jk},
\]

where,
\begin{adjustwidth}{1in}{0pt}
\begin{itemize}[leftmargin=0pt]
  \item[$P_{nijk}$:] Probability that AI output $n$ receives rating category $k$ (e.g., ``good'' rather than ``fair'') on item $i$ from rater $j$.
  \item[$\theta_n$:] Latent quality of output $n$ on the construct of interest (e.g., coherence).
  \item[$\delta_i$:] Difficulty or stringency of item $i$ (e.g., some criteria are harder for outputs to score well on).
  \item[$\rho_j$:] Severity of rater $j$, with larger values corresponding to harsher raters.
  \item[$\tau_jk$:] Threshold between adjacent rating categories $k-1$ and $k$, specific to rater $j$.
\end{itemize}
\end{adjustwidth}

Estimates of $\rho_j$ quantify rater $j$'s level of severity, and estimates of $\theta_n$ quantify AI output $n$'s level of the measured construct, adjusted for item and rater characteristics. There are several ways to specify MFRMs. This particular formulation has been called a \textit{rater-related three-facet partial credit model} \cite{eckes2021measuring}. In this formulation, $\tau_{jk}$ is the threshold for category $k$ specific to rater $j$, capturing the rater’s individual scale usage pattern. Raters who avoid high or low categories exhibit compressed inner thresholds (smaller spacing among $\tau_{jk}$), whereas raters who make full use of the scale exhibit more widely spaced thresholds. This enables the model to distinguish between rater severity ($\rho_j$) and rater centrality or extremity (via patterns in $\tau_{jk}$). To quantify centrality, we estimate the standard deviation of rater threshold parameter estimates, SD($\tau_{jk}$) \cite{eckes2021measuring,myford2004part2}. Larger values of SD($\tau_{jk}$) indicate centrality and smaller values of SD($\tau_{jk}$) indicate extremity. Other model formulations specify explicit parameters for centrality \cite{jin2018new}.

\textbf{What do the model estimates tell us?} The resulting MFRM latent trait estimates can be used as measures of quality at the level of the outputs, and aggregates of the estimates can be used to summarize overall quality of the AI model. Rater parameter estimates can be used to identify raters with aberrant scoring patterns, supporting a flagging system for targeted rater remediation. In practice we might implement a flagging rule to identify the top 5\% of struggling raters using percentiles, separately for each index \cite{stafford2018detecting}. To identify raters with low or high levels on each rater effect index, we would flag the raters below the 2.5th percentile or above the 97.5th percentile. 

\subsection{Data Collection Designs for MFRMs}
For the model to be estimable, data collection must ensure sufficient linkage, or overlap, among raters and responses \cite{wind2021detecting,casabianca2023using,wind2018stabilizing}. We assume an evaluation design in which multiple annotators rate outputs from one or more systems across multiple items (e.g., essays, summaries, dialogues). The resulting data can be structured as a rater × item design (or rater × item × system when comparing systems), where each rating reflects both the latent quality of the system’s output and systematic rater effects.

Collecting multiple ratings per output (e.g., 3–5 annotators per item) provides the replication needed to estimate interrater agreement and to separate true performance from rater error \cite{wind2021detecting}. However, replication alone is insufficient. The design must also include overlap across items and raters. For example, if one group of raters scores only a subset of items that is never scored by other raters, the data become weakly linked, leading to estimation problems. Designs with greater overlap and multiple ratings per output yield more stable estimates, reduce sensitivity to individual rater idiosyncrasies, and are more likely to generalize across evaluation contexts. Extensive discussion of rater linkage is available in the literature and should be reviewed when developing a rating design \cite{wind2019effects,wind2016exploring}.

\section{Application of IRT Rater Models to OpenAI Summarization Data}
\subsection{Data Description}
The OpenAI summarization dataset was made available to support research on training summarization models with reinforcement learning from human feedback (RLHF) \cite{stiennon2020learning}. 
In the study, the RLHF training process involved three steps: (1) collecting human preference judgments between pairs of model-generated summaries, (2) training a reward model to predict human preferences, and (3) using Proximal Policy Optimization (PPO) to train a policy model based on the reward model's scores. (\textit{Policy} refers to the model and training methods.)\footnote{While policy is a term from reinforcement learning referring to a model that selects actions to maximize a reward signal, Stiennon et al. \cite{stiennon2020learning} use it more broadly to refer to any summary-generating method under evaluation, including supervised fine-tuned models, zero-shot pretrained models, and the lead-3 extractive baseline. We use the term in a similar fashion in the current paper.} The researchers trained multiple policies using different approaches, including supervised learning and various RLHF configurations (see Appendix A for a list of policies). To compare these policies, they generated summaries of CNN/Daily Mail (DM) news articles (which were not part of the reward model training data) and collected Likert scale ratings for N=639 summaries. $R$=15 trained raters scored the summaries on 7-point scales (1=poor to 7=excellent) along four quality dimensions: \textit{Content coverage} (how much important information from the original article is covered), \textit{Factual accuracy} (to what degree the statements in the summary are stated in the article), \textit{Coherence} (how easy the summary is to read on its own), and \textit{Overall quality}. Appendix B provides the rubrics for the four items. 

The goal was to compare summaries across different policies/models, including human feedback models, human-written summaries (ref), extractive summaries using the first three sentences (lead3), and other model types.

\subsubsection{Dataset Fitness for MFRM}
The dataset contained 6,312 ratings based on 639 unique articles, summarized by 19 policies which were evaluated by 15 raters on 4 items. Most raters evaluated all 19 summaries for each article they evaluated. The resulting dataset thus contains multiple ratings per summary, with each record linked to the summary text, its generating model or policy, the identity of the rater, and the rater’s scores across the four dimensions. This structure, in which outputs (summaries of CNN/DM articles), items (quality dimensions), and raters jointly contribute to observed scores, makes the dataset well-suited for analysis using MFRM. For many summaries there was only one set of rater's ratings. However, due to the overlap between the raters scoring multiple summaries of the same article and from multiple raters evaluating the same summaries, there is sufficient linkage of raters and source documents to estimate the MFRM.

While a four-item scale is relatively short for IRT model estimation, this limitation is mitigated by two factors: polytomous items (with more than 2 score levels) provide more information per item than dichotomous items, and the model structure is not overly complex.
 
\subsubsection{Interrater Agreement}
\begin{figure}
    \centering
    \includegraphics[width=1\linewidth]{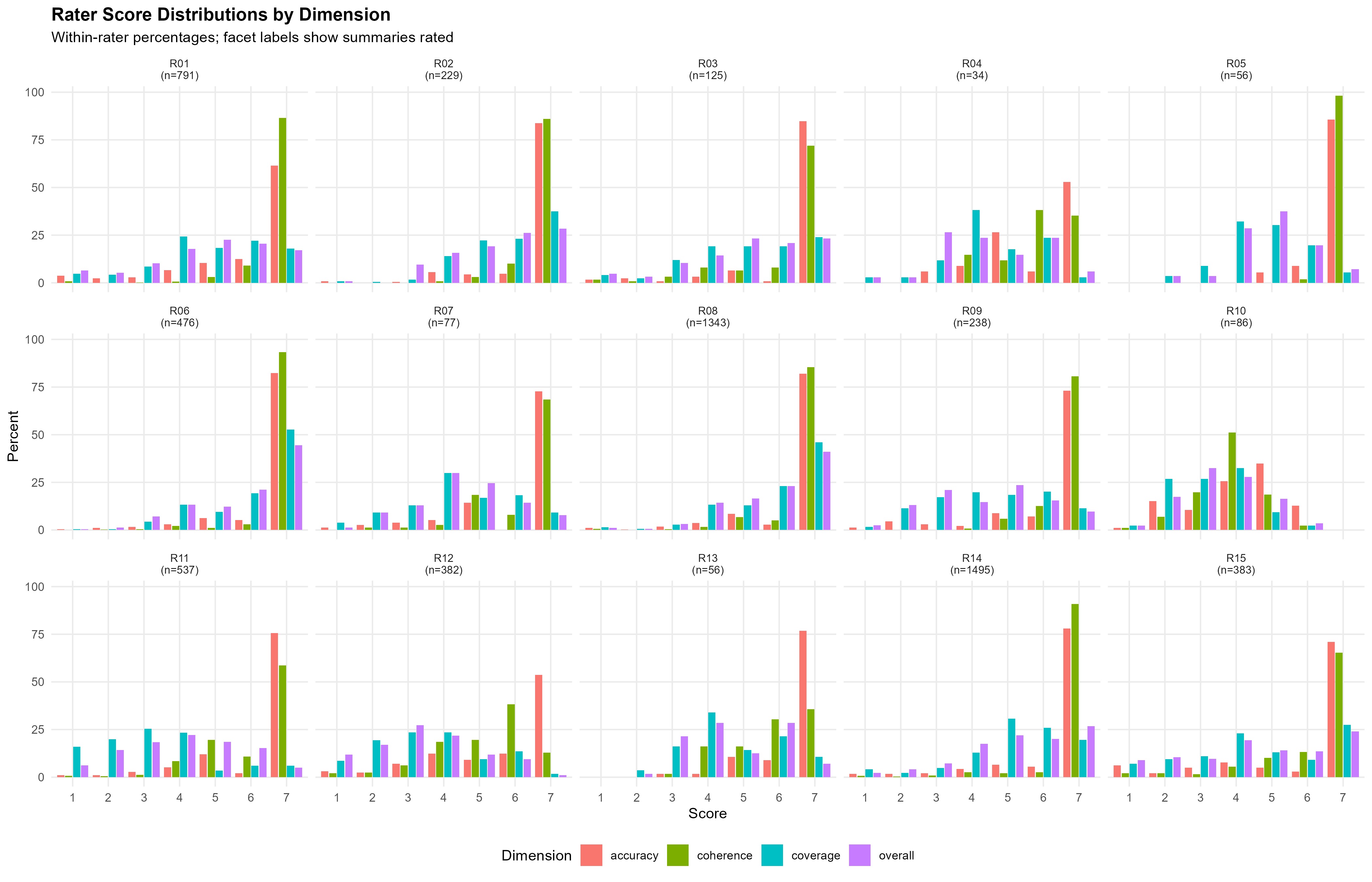}
    \caption{Within-rater score distributions by dimension. Bars show the percentage of ratings assigned to each score (1–7) for \textit{Factual accuracy}, \textit{Coherence}, \textit{Content coverage}, and \textit{Overall quality}, faceted by rater.}
    \label{fig:hist}
\end{figure}

Based on the n = 371 summary*policy combinations that had double ratings from at least two different human raters, the overall quadratic weighted kappa (QWK) and percent exact agreement was poor to moderate, depending on the dimension. QWK ranged from .31 to .50. Agreement was strongest for \textit{Factual accuracy}, which had QWK=.50 (95\% CI [.41, .58]). Exact agreement was relatively high at 68.5\%, while discrepant ratings accounted for 22.6\% of cases, with 77.4\% of ratings falling within one score point and a low mean difference of 0.74. \textit{Coherence} showed weaker agreement, with QWK=.31 (95\% CI [.21, .40]). In contrast, \textit{Content coverage} and \textit{Overall quality} exhibited substantially lower exact agreement but had moderate QWKs; for \textit{Content coverage} and \textit{Overall quality}, QWKs were .48 (95\% CI [.39, .56]) and .47 (95\% CI [.38, .56]), respectively. These  QWK estimates reflect low levels of agreement, however, it is an artifact of the ratings distributions being highly skewed. Figure~\ref{fig:hist} shows a panel of bar charts, one chart for each of the $R$ = 15 raters. Each chart plots the score distribution in percents, with dimensions shown side-by-side in different colors. From this plot, we observe that most raters use the lower end of the score scale sparingly--there is a tendency for raters to assign score of 7, especially for \textit{Factual accuracy} and \textit{Coherence}. This might reflect the true quality of the summaries, or it might reflect the raters' inability to differentiate the score levels. Note that due to small sample size, we did not examine this per policy. However, there are identifiable differences in rater distributions. For example, Raters 04, 10, and 12 do not have the same skewed distribution.

\subsection{Research Questions}
\begin{enumerate}
    \item \textbf{RQ1}: To what extent do individual raters exhibit systematic biases (e.g., severity, leniency, central tendency) as measured by many-facet Rasch measurement (MFRM) rater effect estimates?
    \item \textbf{RQ2}: How do conclusions about policy performance differ when using raw Likert ratings versus MFRM-adjusted scores that account for rater effects?
\end{enumerate}

\subsection{Methodology}
For each policy, we fit two IRT models to the OpenAI summarization evaluation data: the partial credit model (PCM) \cite{masters2016partial}, and the MFRM as specified above. The PCM is an IRT model for polytomous items that excludes rater parameters but does vary the thresholds by item (does not assume each item has the same rating scale).\footnote{Since there are multiple ratings per summary and the scores must be discrete to fit the PCM, we used the rounded mean score across raters, per dimension, for estimation purposes.} We included this model to provide a comparison between examining raw means, a traditional IRT model, and an IRT model incorporating rater parameters. This resulted in multiple sets of model estimates--latent traits which described the quality of the summarization, item parameters for each item (\textit{Coherence}, \textit{Content coverage}, \textit{Factual accuracy}, \textit{Overall quality}), and rater parameters for the MFRM (severity and centrality). Because policies can systematically differ in quality level, variance, and rating difficulty, we explored a version of the MFRM that includes policy as a facet but that was not estimable due to identification issues. While not discussed here for brevity, IRT models hold several assumptions which we explored and report in Appendix C.

\subsection{Results}
\subsubsection{RQ 1: Analysis of Raters}
The first research question asked to what extent the 15 raters exhibited rater effects. Figure~\ref{fig:rater-level-plots} shows a panel of plots with the policy-based rater centrality estimates (y-axis) and rater severity estimates (x-axis). Each plot provides a rater's estimates for all policies. Note that the axis limits vary across plots and the points are sized based on the sample size. There is notable variability in levels of centrality and severity across raters, and within raters, by policy. 

Most raters have severity estimates near or around 0. However, there are some raters, including R02, R08, and R06, who consistently have negative estimates, indicating they are more lenient in their scoring relative to other raters. Raters with mostly positive estimates, including Raters R04, R10, and 12, tend to be relatively severe in their scoring behavior. 

There were several raters with varying levels of centrality across policies. Many of these estimates were based on smaller samples which may indicate these are noisy estimates of the rater thresholds (and their standard deviations). Rater 10 had very large SDs for many policies, indicative of central tendency (scoring in the middle of the scale). 

\begin{figure}
    \centering
    \includegraphics[width=.8\linewidth]{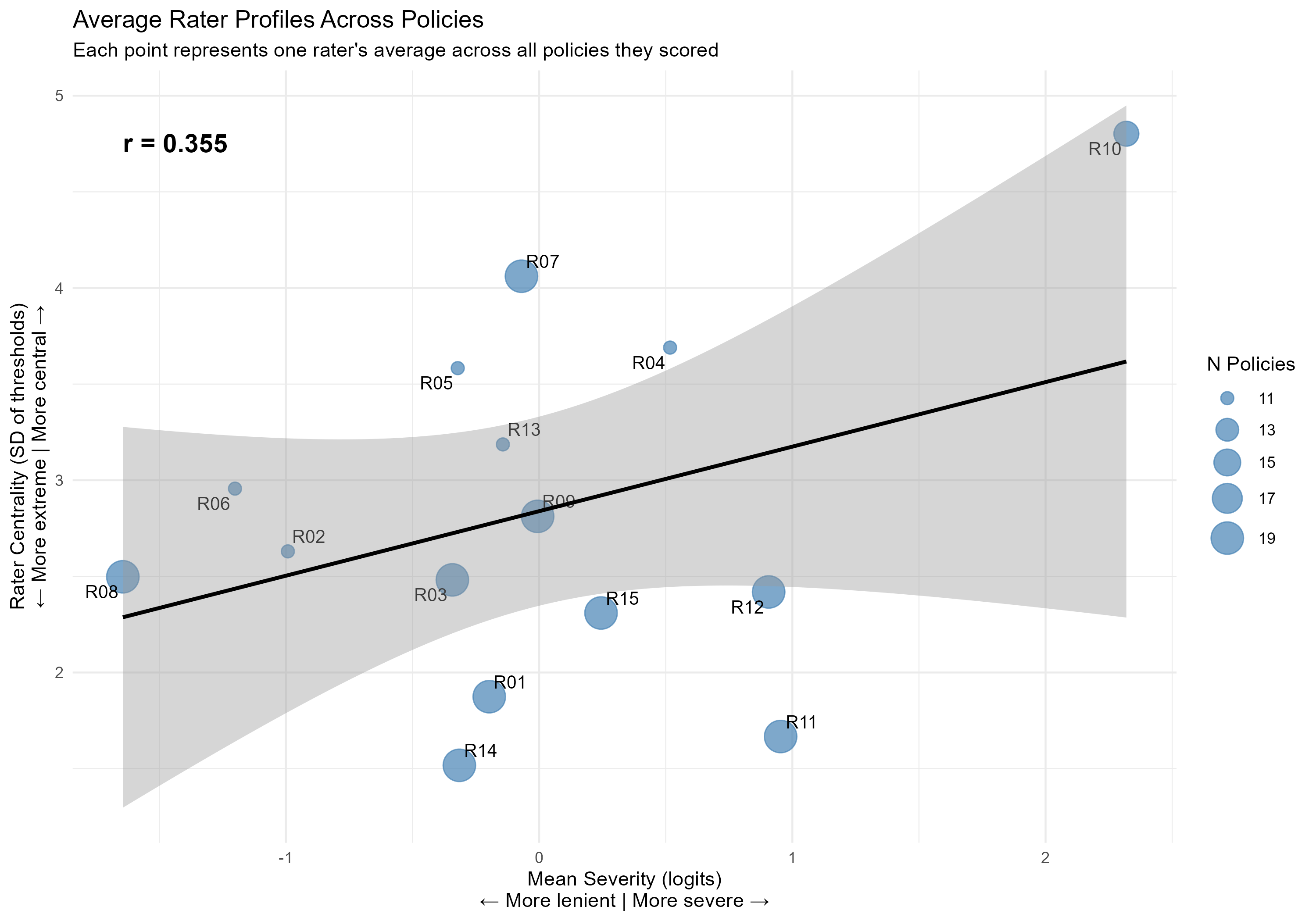}
    \caption{Average rater profiles across policies. Each point represents a rater’s mean MFRM severity (logits) and mean centrality (standard deviation of category thresholds) averaged over all policies scored. Point size reflects the number of policies rated; the solid line shows the linear association with 95\%\ confidence band (Pearson r shown).}
    \label{fig:average-rater-profiles}
\end{figure}

Without applying a formal rule, it is clear that Rater 10 should be examined more closely. Figure~\ref{fig:average-rater-profiles} plots these estimates on one figure for comparison across policies, which shows R10's aberrance.

\subsubsection{RQ 2: Raw vs. Adjusted Policy Rankings}
To compare how the policies ranked after adjusting for rater effects, we plotted the mean ratings (across four items) per policy as a function of model size in Figure~\ref{fig:quality_clusters_combined}(a) and plotted the mean latent trait estimates from the PCMs and MFRMS in (b) and (c). It is important to note that the scales are not comparable across panels. That is, we cannot examine differences in the trait estimates, but we can compare the relative rankings. 

The plot in (a) confirms the same findings from Steinnon \cite{stiennon2020learning}--the T5 model policy produced summaries with the highest average quality, falling in between the baseline reference summaries written by humans and Lead3 (first three sentences of source CNN/DM news articles). The Supervised CNN/DM

\begin{landscape}
\centering
\begin{figure}
    \centering
    \includegraphics[width=.8\linewidth]{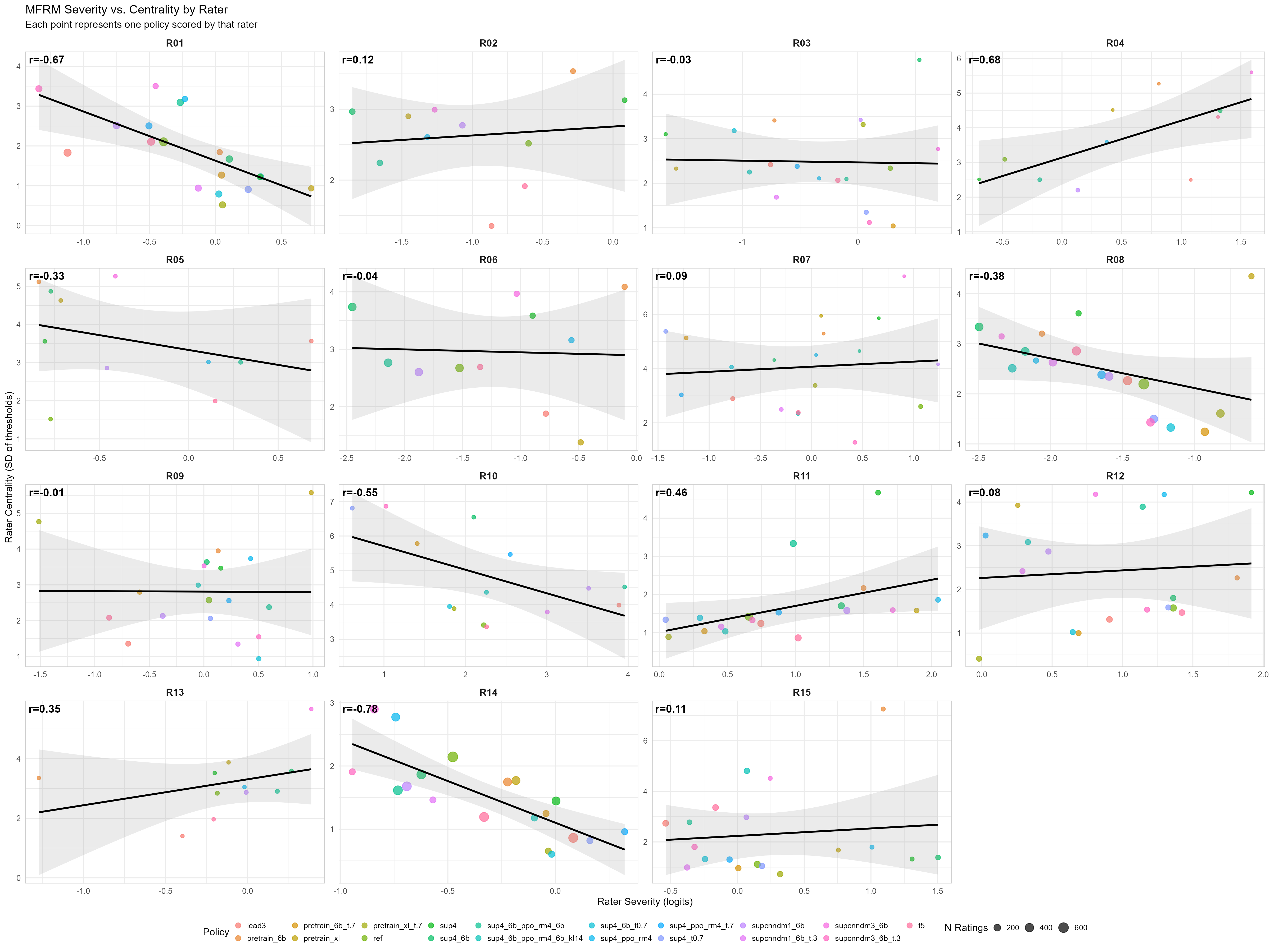}
    \caption{Array of scatterplots showing MFRM severity (logits) versus centrality (SD of thresholds), by rater. Each panel shows a single rater; each point represents a policy scored by that rater. Solid lines indicate linear trends with 95\%\ confidence bands.}
    \label{fig:rater-level-plots}
\end{figure}
\end{landscape}

model (\texttt{supcnndm1\_6b}) which was actually \textit{trained} on CNN/DM reference summaries, was in second place. Following that was one of the human feedback models \texttt{sup4\_6b\_ppo\_rm4\_6b}. 

In the PCM results shown in panel (b), one of the human feedback models \texttt{sup4\_6b\_ppo\_rm4\_6b} has outranked \texttt{supcnndm1\_6b} and sits just below T5. Moving to panel (c) we observe a more drastic shift in rankings on the latent trait scale such that now, two human feedback models, \texttt{sup4\_6b\_ppo\_rm4\_6b} and \texttt{sup4\_ppo\_rm4}, outperform all other models and the human-generated reference summaries. Following these two models are T5 and \texttt{supcnndm1\_6b}. 

Once rater severity and centrality are modeled in the MFRM, the human feedback model results showed highest overall summary quality. Other policies, for example \texttt{pretrain\_6b} and \texttt{pretrain\_xl} remained consistently low across all metrics, signaling evidence of weaker performance. The figure demonstrates that raw-score rankings misrepresent the relative standing of models. Some policies appear strong only because they were scored by lenient raters, while others appear weak due to unusually harsh or central raters.

Figure~\ref{fig:rater-effects-bypolicy} shows the rater effects estimates by policy. When reviewing \texttt{sup4\_6b\_ppo\_rm4\_6b} it is clear that the raters who contributed to the latent trait estimates were scoring severely and with central tendency (R10). On the other hand, the group of pretrained models appearing in the top row of the panel in Figure~\ref{fig:rater-effects-bypolicy} display points with severity estimates that are fairly spread about 0.

\section{Conclusion}
This short paper reports some empirical results from a larger line of inquiry focused on understanding how latent variable models can be leveraged in the human data annotation pipeline. From these results, we can see that human errors can impact AI evaluation. In practical terms, relying on raw means might lead us to select suboptimal systems and overlook policies that perform well once evaluated on a level playing field.

\clearpage
\begin{landscape}
\begin{figure}
\centering
\includegraphics[width=1\linewidth]
{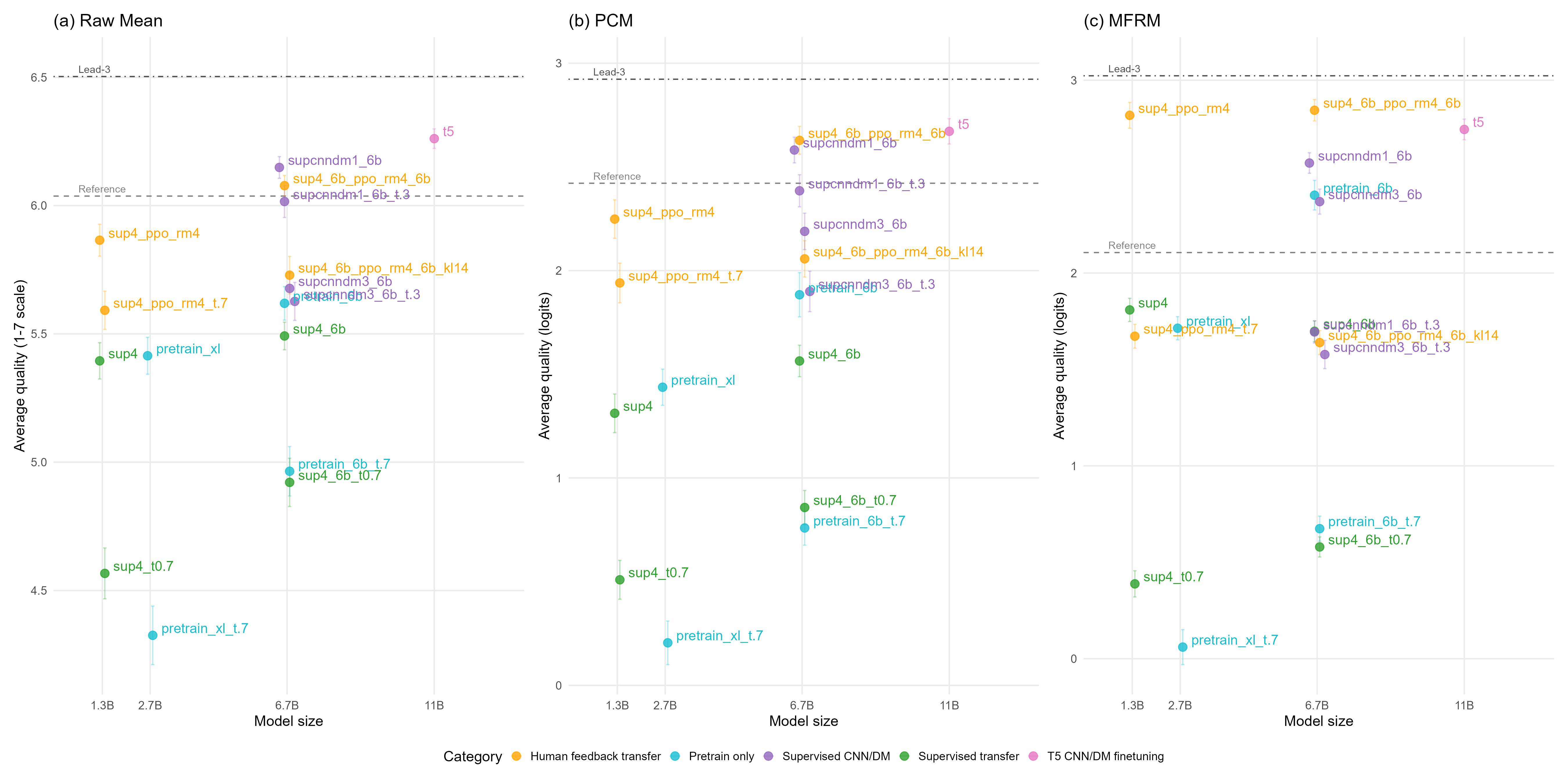}
\caption{Policy-level summary quality based on raw scores and IRT model estimates. Average quality estimates are plotted by model size and color coded by training method. Panel (a) shows mean human ratings across four items on the 7-point scale, while panel (b) shows the PCM estimates, and (c) shows corresponding MFRM quality estimates ($\theta$, logits). Points represent model variants, grouped by model size. Dashed horizontal reference lines mark human-written summaries and the Lead-3 baseline. Note: While the value of the estimates are not directly comparable, the differences in rank ordering are meaningful. }
\label{fig:quality_clusters_combined}
\end{figure}

\clearpage

\begin{figure}
    \centering
    \includegraphics[width=.9\linewidth]{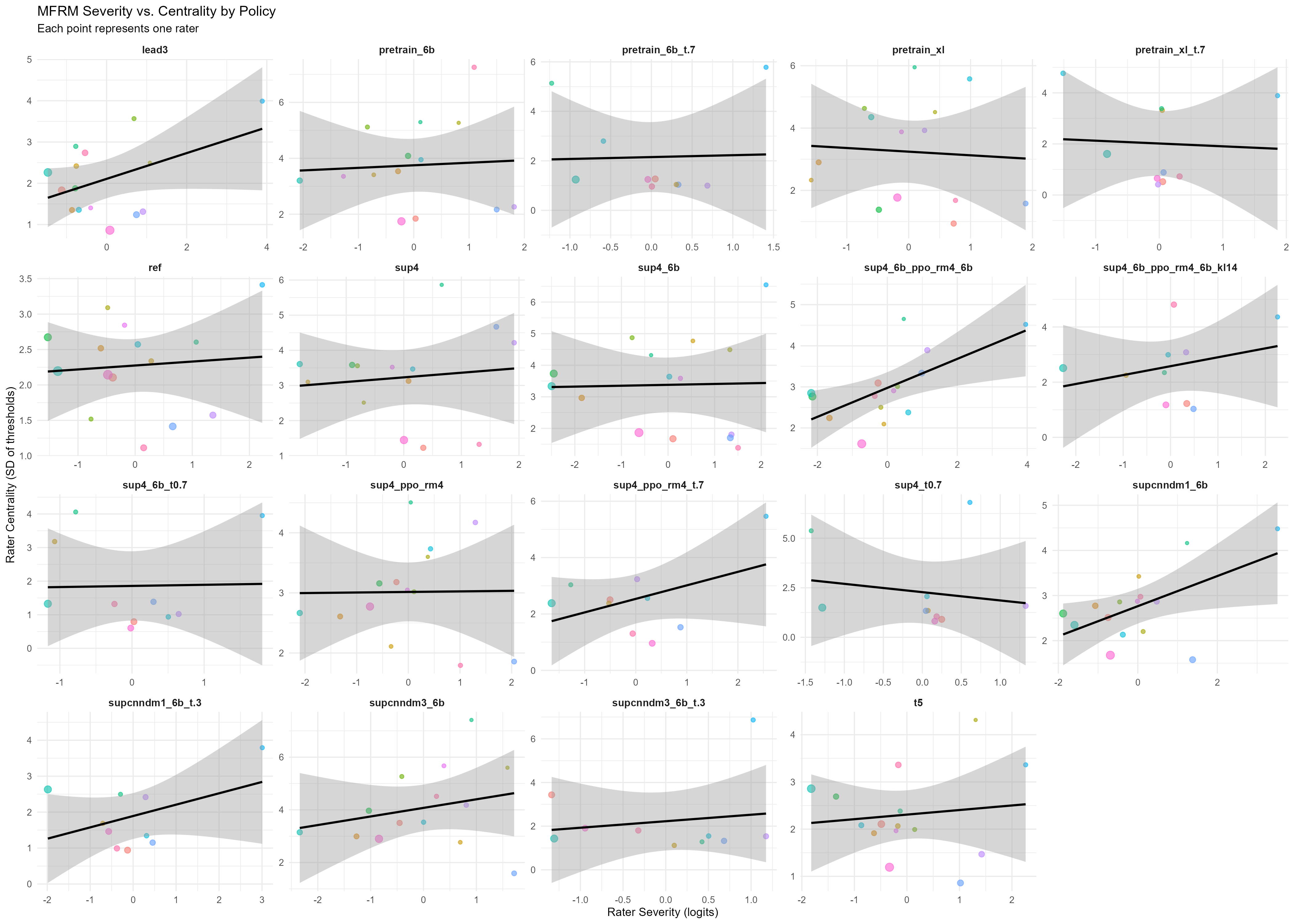}
    \caption{Rater severity and centrality by policy. Each panel shows one policy. Points represent individual raters. The line shows the overall trend for each policy.}
    \label{fig:rater-effects-bypolicy}
\end{figure}

\end{landscape}

This methodology may be used in different ways. One use case is to identify raters with large errors and use the information diagnostically to give feedback for rater remediation. Often in human annotation projects, only superficial tracking of raters may be used, such as basic inter-rater reliability metrics. These metrics are not highly actionable. Estimating various rater effects can assist scoring leaders in targeted training remediation efforts. The other use case is to use the IRT model estimates to make decisions about the model, not the raw ratings data.

We use the MFRM here because it is a relatively simple model that is widely used to quantify common rater effects---there are many alternative rater models that could be explored. It is important to note that IRT rater models make several assumptions which must be checked. Assumption checks indicated approximate unidimensionality and no evidence of problematic local dependence, though conclusions are limited by the short scale length. The results in this paper should be viewed as a proof-of-concept rather than a definitive measurement analysis. An analysis of a larger sample of ratings with more dimensions related to summary quality would be important to corroborate these findings. Importantly, a central takeaway for those incorporating IRT rater models into their pipelines is the importance of conducting meta-evaluations of the scales and rubrics used in both human and LLM-based evaluations to ensure alignment with—and adequate representation of—the intended construct.\cite{casabiancapsych}.

\bibliography{3sample-ceur}

@String{Chelsea = "Chelsea" }

@article{wind2021detecting,
  title={Detecting rater biases in sparse rater-mediated assessment networks},
  author={Wind, Stefanie A and Ge, Yuan},
  journal={Educational and Psychological Measurement},
  volume={81},
  number={5},
  pages={996--1022},
  year={2021},
  publisher={SAGE Publications Sage CA: Los Angeles, CA}
}

@article{mcclellan2010constructed,
  title={Constructed-response scoring—Doing it right},
  author={McClellan, Catherine A},
  journal={R\&D Connections},
  volume={13},
  pages={1--7},
  year={2010}
}

@book{lord2012applications,
  title={Applications of item response theory to practical testing problems},
  author={Lord, Frederic M},
  year={2012},
  publisher={Routledge}
}

@article{wolfe2012application,
  title={Application of latent trait models to identifying substantively interesting raters},
  author={Wolfe, Edward W and McVay, Aaron},
  journal={Educational Measurement: Issues and Practice},
  volume={31},
  number={3},
  pages={31--37},
  year={2012},
  publisher={Wiley Online Library}
}

@article{wind2016exploring,
  title={Exploring the effects of rater linking designs and rater fit on achievement estimates within the context of music performance assessments},
  author={Wind, Stefanie A and Engelhard Jr, George and Wesolowski, Brian},
  journal={Educational Assessment},
  volume={21},
  number={4},
  pages={278--299},
  year={2016},
  publisher={Taylor \& Francis}
}

@article{wind2019effects,
  title={The effects of incomplete rating designs in combination with rater effects},
  author={Wind, Stefanie A and Jones, Eli},
  journal={Journal of Educational Measurement},
  volume={56},
  number={1},
  pages={76--100},
  year={2019},
  publisher={Wiley Online Library}
}

@article{robitzsch2018item,
  title={Item response models for human ratings: Overview, estimation methods, and implementation in R},
  author={Robitzsch, Alexander and Steinfeld, Jan},
  journal={Psychological Test and Assessment Modeling},
  volume={60},
  number={1},
  pages={101--138},
  year={2018},
  publisher={PABST Science Publishers}
}

@article{maline2025unmasking,
  title={Unmasking survey fraud: investigating data quality issues in an MTurk sample},
  author={Maline, Marissa N and Polonijo, Andrea N},
  journal={International Journal of Social Research Methodology},
  pages={1--8},
  year={2025},
  publisher={Taylor \& Francis}
}

@article{jin2018new,
  title={A new facets model for rater's centrality/extremity response style},
  author={Jin, Kuan-Yu and Wang, Wen-Chung},
  journal={Journal of Educational Measurement},
  volume={55},
  number={4},
  pages={543--563},
  year={2018},
  publisher={Wiley Online Library}
}

@article{wind2018stabilizing,
  title={The stabilizing influences of linking set size and model--data fit in sparse rater-mediated assessment networks},
  author={Wind, Stefanie A and Jones, Eli},
  journal={Educational and Psychological Measurement},
  volume={78},
  number={4},
  pages={679--707},
  year={2018},
  publisher={Sage Publications Sage CA: Los Angeles, CA}
}

@article{casabianca2023using,
  title={Using linkage sets to improve connectedness in rater response model estimation},
  author={Casabianca, Jodi M and Donoghue, John R and Shin, Hyo Jeong and Chao, Szu-Fu and Choi, Ikkyu},
  journal={Journal of Educational Measurement},
  volume={60},
  number={3},
  pages={428--454},
  year={2023},
  publisher={Wiley Online Library}
}

@article{eckes2021measuring,
  title={Measuring rater centrality effects in writing assessment: A Bayesian facets modeling approach},
  author={Eckes, Thomas and Jin, Kuan-Yu},
  journal={Psychological Test and Assessment Modeling},
  volume={63},
  number={1},
  pages={65--94},
  year={2021},
  publisher={PABST Science Publishers}
}

@Incollection{casabianca2022rater,
  author =       "Casabianca, Jodi M",
  title =        "Rater effects modeling",
  booktitle =    "International Encyclopedia of Education",
  publisher =    "Elsevier",
  volume =       "4e",
  year =         "2022",
    pages={547--581},
  chapter =      "",
  editor =       "Daniel F. McCaffrey"}

@article{nieto2019accounting,
  title={Accounting for rater effects with the hierarchical rater model framework when scoring simple structured constructed response tests},
  author={Nieto, Ricardo and Casabianca, Jodi M},
  journal={Journal of Educational Measurement},
  volume={56},
  number={3},
  pages={547--581},
  year={2019},
  publisher={Wiley Online Library}
}

@article{mccaffrey2022best,
  title={Best Practices for Constructed-Response Scoring},
  author={McCaffrey, Daniel F and Casabianca, Jodi M and Ricker-Pedley, Kathryn L and Lawless, Ren{\'e} R and Wendler, Cathy},
  journal={ETS Research Report Series},
  volume={2022},
  number={1},
  pages={1--58},
  year={2022},
  publisher={Wiley Online Library}
}

@article{likert1932technique,
  title={A technique for the measurement of attitudes},
  author={Likert, Rensis},
  journal={Archives of Psychology},
  volume={22},
  number={140},
  pages={1--55},
  year={1932}
}

@article{myford2003part1,
  author  = {Myford, Carol M. and Wolfe, Edward W.},
  title   = {Detecting and Measuring Rater Effects Using Many-Facet Rasch Measurement: Part I},
  journal = {Journal of Applied Measurement},
  volume  = {4},
  number  = {4},
  pages   = {386--422},
  year    = {2003}
}

@article{myford2004part2,
  author  = {Myford, Carol M. and Wolfe, Edward W.},
  title   = {Detecting and Measuring Rater Effects Using Many-Facet Rasch Measurement: Part II},
  journal = {Journal of Applied Measurement},
  volume  = {5},
  number  = {2},
  pages   = {189--227},
  year    = {2004}
}

@incollection{masters2016partial,
  title={Partial credit model},
  author={Masters, Geoff N},
  booktitle={Handbook of item response theory},
  pages={109--126},
  year={2016},
  publisher={Chapman and Hall/CRC}
}

@inproceedings{hardy2025all,
  title={All that Glitters: Techniques for Evaluations with Unreliable Model and Human Annotations},
  author={Hardy, Michael},
  booktitle={Findings of the Association for Computational Linguistics: NAACL 2025},
  pages={2250--2278},
  year={2025}
}

@article{casabiancapsych,
  title={Psychometrics is all you need},
  author={Casabianca, Jodi M},
  journal={EdArXiv preprint \url{https://osf.io/preprints/edarxiv/7w6pz\_v1}},
  year={2025}
}

@article{fabbri2021summeval,
  title={Summeval: Re-evaluating summarization evaluation},
  author={Fabbri, Alexander R and Kry{\'s}ci{\'n}ski, Wojciech and McCann, Bryan and Xiong, Caiming and Socher, Richard and Radev, Dragomir},
  journal={Transactions of the Association for Computational Linguistics},
  volume={9},
  pages={391--409},
  year={2021},
  publisher={MIT Press One Rogers Street, Cambridge, MA 02142-1209, USA journals-info~…}
}

@article{stiennon2020learning,
  title={Learning to summarize with human feedback},
  author={Stiennon, Nisan and Ouyang, Long and Wu, Jeffrey and Ziegler, Daniel and Lowe, Ryan and Voss, Chelsea and Radford, Alec and Amodei, Dario and Christiano, Paul F},
  journal={Advances in neural information processing systems},
  volume={33},
  pages={3008--3021},
  year={2020}
}

@article{stafford2018detecting,
  title={Detecting Rater Effects under Rating Designs with Varying Levels of Missingness.},
  author={Stafford, Rose E and Wolfe, Edward W and Casablanca, Jodi M and Song, Tian},
  journal={Journal of Applied Measurement},
  volume={19},
  number={3},
  pages={243--257},
  year={2018}
}

@phdthesis{linacre1989many,
  title={Many-faceted Rasch measurement},
  author={Linacre, John Michael},
  year={1989},
  school={The University of Chicago}
}

\clearpage
\FloatBarrier
\appendix

\begin{landscape}
\section{Policies from OpenAI Summarization Dataset}
\begin{table}[h]
\centering
\small
\begin{tabular}{lllllp{4cm}}
\hline
\textbf{Policy Name} & \textbf{Description} & \textbf{Training Method} & \textbf{Model Size} & \textbf{Temp.} & \textbf{Dataset} \\
\hline

ref & Reference summaries & Human-written & N/A & N/A & CNN/DM \\
lead3 & Lead-3 baseline & Extractive (first 3 sentences) & N/A & N/A & CNN/DM \\

pretrain\_xl & Pretrained only & Zero-shot & $\sim$3B & 0 & Pretrain \\
pretrain\_xl\_t.7 & Pretrained only & Zero-shot & $\sim$3B & 0.7 & Pretrain \\
pretrain\_6b & Pretrained only & Zero-shot & 6.7B & 0 & Pretrain \\
pretrain\_6b\_t.7 & Pretrained only & Zero-shot & 6.7B & 0.7 & Pretrain \\

sup4 & Supervised TL;DR & Supervised learning & 1.3B & 0 & TL;DR $\rightarrow$ CNN/DM \\
sup4\_t0.7 & Supervised TL;DR & Supervised learning & 1.3B & 0.7 & TL;DR $\rightarrow$ CNN/DM \\
supcnndm1\_6b & Supervised CNN/DM & Supervised learning & 6.7B & 0 & CNN/DM \\
supcnndm3\_6b & Supervised CNN/DM v3 & Supervised learning & 6.7B & 0 & CNN/DM \\
sup4\_6b & Supervised TL;DR & Supervised learning & 6.7B & 0 & TL;DR $\rightarrow$ CNN/DM \\
supcnndm1\_6b\_t.3 & Supervised CNN/DM & Supervised learning & 6.7B & 0.3 & CNN/DM \\
supcnndm3\_6b\_t.3 & Supervised CNN/DM v3 & Supervised learning & 6.7B & 0.3 & CNN/DM \\
sup4\_6b\_t0.7 & Supervised TL;DR & Supervised learning & 6.7B & 0.7 & TL;DR $\rightarrow$ CNN/DM \\
t5 & T5 model & Supervised (encoder-decoder) & 11B & Beam & CNN/DM \\

sup4\_ppo\_rm4 & Human feedback (RL) & PPO with RM4 & 1.3B & 0 & TL;DR $\rightarrow$ CNN/DM \\
sup4\_ppo\_rm4\_t.7 & Human feedback (RL) & PPO with RM4 & 1.3B & 0.7 & TL;DR $\rightarrow$ CNN/DM \\
sup4\_6b\_ppo\_rm4\_6b & Human feedback (RL) & PPO with RM4 & 6.7B & 0 & TL;DR $\rightarrow$ CNN/DM \\
sup4\_6b\_ppo\_rm4\_6b\_kl14 & Human feedback (RL) & PPO (KL=14) & 6.7B & 0 & TL;DR $\rightarrow$ CNN/DM \\

\hline
\end{tabular}
\caption{Summary of policies evaluated on CNN/DM (\cite{stiennon2020learning}). Key: \textbf{sup} = supervised learning, \textbf{ppo} = Proximal Policy Optimization (RL), \textbf{rm} = reward model, \textbf{cnndm} = CNN/DailyMail dataset, \textbf{t.X} = temperature parameter, \textbf{kl} = KL divergence penalty.}
\label{tab:policies}
\end{table}
\end{landscape}

\section{Instructions Given to Labelers for Evaluating Summaries}

\subsection{Four axes of quality}

\subsubsection{Coherence}
For this axis, answer the question “how coherent is the summary on its own?” A summary is coherent if, when read by itself, it’s easy to understand and free of English errors. A summary is not coherent if it’s difficult to understand what the summary is trying to say. Generally, it’s more
important that the summary is understandable than it being free of grammar errors.

Rubric:

\begin{itemize}
\item Score of 1: The summary is impossible to understand.
\item Score of 4: The summary has mistakes or confusing phrasing that make it a bit hard to understand.
\item Score of 7: The summary is perfectly clear.
\end{itemize}

\subsubsection{Accuracy}
For this axis, answer the question “does the factual information in the summary accurately match the post?” A summary is accurate if it doesn’t say things that aren’t in the article, it doesn’t mix up people, and generally is not misleading. If the summary says anything at all that is not mentioned in the post or contradicts something in the post, it should be given a maximum score of 5. 

Rubric:
\begin{itemize}
\item Score of 1: The summary is completely wrong, made up, or exactly contradicts what is written in the post.
\item Score of 4: The summary says at least one substantial thing that is not mentioned in the post, or that contradicts something in the post.
\item Score of 5: The summary says anything, no matter how small, that is not mentioned in the post, or that contradicts something in the post.
\item Score of 7: The summary has no incorrect statements or misleading implications.
\end{itemize}

\subsubsection{Coverage}
For this axis, answer the question “how well does the summary cover the important information in the post?” A summary has good coverage if it mentions the main information from the post that’s important to understand the situation described in the post. A summary has poor coverage if someone reading only the summary would be missing several important pieces of information about the situation in the post. A summary with good coverage should also match the purpose of the original post (e.g. to ask for advice).

Rubric:
\begin{itemize}
\item Score of 1: The summary contains no information relevant to the post.
\item Score of 4: The summary is missing at least 1 important piece of information required to understand the situation.
\item Score of 7: The summary covers all of the important information required to understand the situation.
\end{itemize}

\subsubsection{Overall quality}
For this axis, answer the question “how good is the summary overall at representing the post?” This can encompass all of the above axes of quality, as well as others you feel are important. If it’s hard to find ways to make the summary better, give the summary a high score. If there are lots of different ways the summary can be made better, give the summary a low score.

Rubric:
\begin{itemize}
\item Score of 1: The summary is terrible.
\item Score of 4: The summary is an okay representation of the post, but could be significantly improved.
\item Score of 7: The summary is an excellent representation of the post.

\end{itemize}

\clearpage
\section{Checks of IRT Assumptions}
IRT models assume approximate unidimensionality and local independence of item responses. To evaluate whether the four-item scale met core IRT assumptions, we examined unidimensionality and local independence using standard factor-analytic and residual diagnostics, recognizing the limitations imposed by a short four-item scale. Across 19 policies, assumption checks supported approximate unidimensionality and limited local dependence, with only one policy showing a notable item-pair dependence and no policies exhibiting systematic violations of IRT assumptions.

\subsection{Unidimensionality}
Across the 19 policies, the four rating dimensions showed strong internal consistency ($\alpha$: mean = 0.85, median = 0.86, range = 0.77–0.92) and the one-factor CFA showed near-perfect incremental fit (CFI: mean = 0.9996; TLI: mean = 0.9988). Absolute fit indices were more variable (RMSEA: median = 0.047, max = 0.155; SRMR: median = 0.042, max = 0.118), which is expected with a very short scale (4 items; df = 2). 

\subsection{Local Independence}
Local independence checks were generally acceptable: the maximum CFA residual correlation was modest on average (median = 0.09) with a small number of policies exceeding 0.20, and Yen’s Q3 residual correlations were typically negative (mean $\approx - 0.21$) with few notable positive residual associations (max positive Q3 up to 0.35 in an isolated condition). Overall, these diagnostics support treating the scale as approximately unidimensional with limited local dependence for proof-of-concept rater-model analyses, while recognizing that results are constrained by the short scale length.

\end{document}